\documentclass[10pt,twocolumn,letterpaper]{article}

\usepackage[pagenumbers]{cvpr} 
\usepackage[accsupp]{axessibility}
\usepackage{adjustbox}
\usepackage{booktabs}
\usepackage{pifont}
\usepackage{multicol}
\usepackage{multirow}
\usepackage{threeparttable}

\usepackage[dvipsnames]{xcolor}

\definecolor{cvprblue}{rgb}{0.21,0.49,0.74}
\usepackage[pagebackref,breaklinks,colorlinks,citecolor=cvprblue]{hyperref}

\def\paperID{19} 
\def\confName{CVPR}
\def\confYear{2024}

\title{TattTRN: Template Reconstruction Network for Tattoo Retrieval}

\author{Lazaro Janier Gonzalez-Soler\textsuperscript{1},
Maciej Salwowski\textsuperscript{2},
Christian Rathgeb\textsuperscript{1} and
Daniel Fischer\textsuperscript{1}, \\
\textsuperscript{1}da/sec - Biometrics and Security Research Group, Darmstadt, Germany\\
{\tt\small \{lazaro-janier.gonzalez-soler,christian.rathgeb,daniel.fischer\}@h-da.de}\\
\textsuperscript{2} Faculty of Computer Science, Technical University of Denmark, Denmark\\
{\tt\small s223525@student.dtu.dk}\\
}

\begin{document}
\maketitle

\newcommand{\xmark}{\text{\ding{55}}}

\begin{abstract}
Tattoos have been used effectively as soft biometrics to assist law enforcement in the identification of offenders and victims, as they contain discriminative information, and are a useful indicator to locate members of a criminal gang or organisation. Due to various privacy issues in the acquisition of images containing tattoos, only a limited number of databases exists. This lack of databases has delayed the development of new methods to effectively retrieve a potential suspect's tattoo images from a candidate gallery. To mitigate this issue, in our work, we use an unsupervised generative approach to create a balanced database consisting of 28,550 semi-synthetic images with tattooed subjects from 571 tattoo categories. Further, we introduce a novel Tattoo Template Reconstruction Network (TattTRN), which learns to map the input tattoo sample to its respective tattoo template to enhance the distinguishing attributes of the final feature embedding. Experimental results with real data, \ie WebTattoo and BIVTatt databases, demonstrate the soundness of the presented approach: an accuracy of up to 99\% is achieved for checking at most the first 20 entries of the candidate list.\footnote{\url{https://github.com/ljsoler/TattTRN}}
\end{abstract}

\section{Introduction}
\label{sec:intro}

The use of tattoos as soft biometrics to assist law enforcement in identifying suspects has steadily grown along with their popularity in society. In 2015, The National Institute of Standards and Technology (NIST) (Tatt-C)~\cite{Ngan-Tatt-C-ISBA-2015} reported that one-fifth of US adults have at least one tattoo, making the US population the third most tattooed in the world, after Italy and Sweden. This trend was constantly expanding, as evidenced by a survey conducted in 2021\footnote{\url{https://www.statista.com/statistics/259598/share-of-americans-with-at-least-one-tattoo/}}, which revealed that 26\% of Americans have at least one tattoo. In contrast to biometric characteristics, \eg  fingerprints and faces, tattoos cannot be used to directly establish the identity of a subject. However, tattoos, unlike other soft biometrics such as gender, age or race, contain more discriminative information to support the identification of individuals and are a useful indicator to track members of a criminal gang or organisation~\cite{Ngan-Tatt-C-ISBA-2015}. Therefore, tattoo recognition represents an area of interest for forensic investigators, which motivates the development of automated image-based tattoo retrieval techniques~\cite{Mun-TattooMeaning-2012}.  

\begin{table}[!t]
\caption{Overview of available tattoo databases.}
\label{tab:available_databases}\vspace{-0.2cm}
\begin{adjustbox}{width=\columnwidth,center}
\begin{threeparttable}
    \begin{tabular}{r| c c c c} \toprule
      \textbf{Database}                                     &\textbf{\#Categories} &\textbf{\#Samples}  & \textbf{Public}    & \textbf{Semi-synthetic}  \\\midrule 
HDA-STD~\cite{GonzalezSoler-SemiSyntheticTattoo-IWBF-2023}  &        -           &    5,500           &          Yes       &    \checkmark \\
      WebTattoo~\cite{Han-TattooImgSearch-PAMI-2019}        &      400$^*$       &    1,400           &          Yes       &    $\xmark$\\
      DeMSI~\cite{Hrkac-Tattoo-DeMSI-2016}                  &       -            &      890           &          Yes       &     $\xmark$ \\
      Tatt-C~\cite{Ngan-Tatt-C-ISBA-2015}                   &      157$^\dagger$ &      215           &          No        &    $\xmark$\\
      BIVTatt~\cite{Nicolas-DeepGenericFeatures-CIARP-2019} &      210           &    4,200           &          Yes       &    $\xmark$\\
      PinTatt~\cite{Nicolas-DeepGenericFeatures-CIARP-2019} &      160           &    454             &          No       &    $\xmark$\\
      NTU-Tattoo-V1~\cite{Xu-NTUTattooV1-ICB-2016}          &       -            &    10,000          &          Yes       &    $\xmark$ \\
      \textbf{Ours}                                         &   \textbf{571}     & \textbf{28,550}    &  \textbf{Yes}      &    \checkmark \\
      \bottomrule
    \end{tabular}
    \begin{tablenotes}\footnotesize
		\item[*] Publicly available training set.
            \item[$\dagger$] Number of categories reported for the identification case.
    \end{tablenotes}
    \end{threeparttable}
\end{adjustbox}
\end{table} 

In the context of tattoo retrieval, the NIST Tatt-C and Tatt-E challenges advanced the development of tattoo detection and identification systems for real-world application scenarios~\cite{Ngan-TattC-Outcomes-2016}. Earlier tattoo image retrieval practices were based on keyword or metadata matching, where law enforcement agencies typically followed the ANSI/NIST-ITL 1-2000 standard and assigned a single keyword to each tattoo image in the database~\cite{Han-TattooImgSearch-PAMI-2019}. The limitations of keyword-based systems (\eg, limited and insufficient vocabulary to describe different tattoo patterns and inconsistent labels) led to the development of techniques that represented tattoos as handcrafted descriptors capturing the texture, appearance, or colour of the tattoos~\cite{Li-SIFTImageRetrieval-2009,Lee-ImageRetrievalForensics-2011,Han-TattooIdentSketch-ICB-2013}. 

\begin{figure}[!t]
    \centering
    \includegraphics[width=\linewidth]{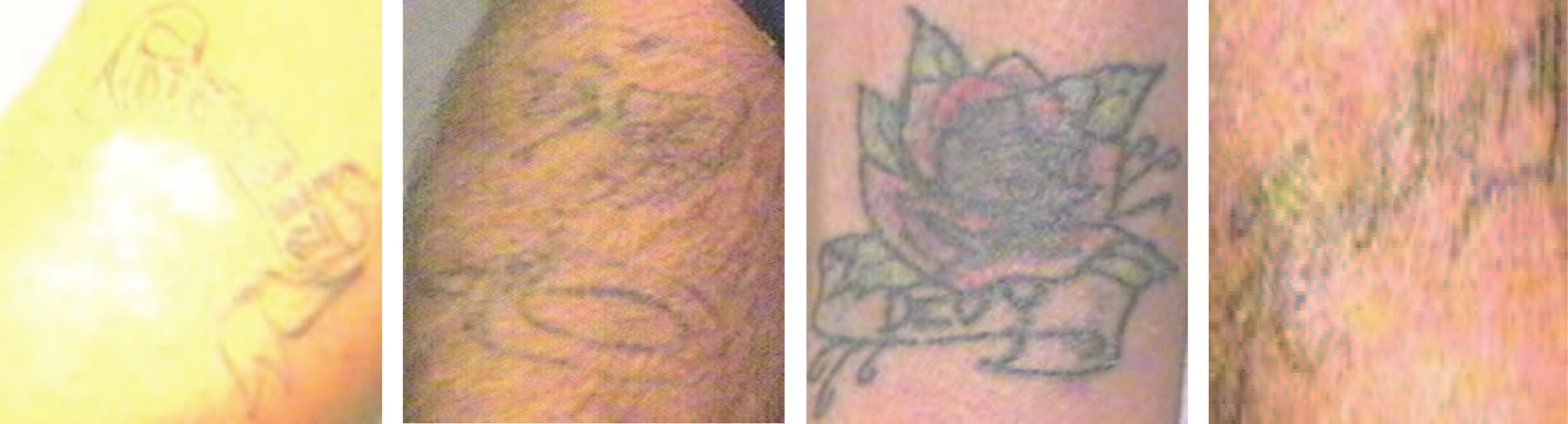}
    \caption{Examples of tattoos from WebTattoo~\cite{Han-TattooImgSearch-PAMI-2019} representing a challenge for state-of-the-art solutions.}
    \label{fig:tattoo_issues}
\end{figure}

With the introduction and success of deep neural networks (DNNs) in various pattern recognition and computer vision applications, some recent research efforts have been devoted to tattoo localisation~\cite{Sun-TattooDetectionLocalisation-ICPR-2016}, detection~\cite{Sun-TattooDetectionLocalisation-ICPR-2016,Da-TransferLearningDetection-2021} or segmentation~\cite{GonzalezSoler-SemiSyntheticTattoo-IWBF-2023}. Moreover, a few techniques have focused on tattoo retrieval or identification through the representation of tattoos as compact binary~\cite{Han-TattooImgSearch-PAMI-2019} or floating-point embeddings~\cite{Di-DeepTattoo-CVPRW-2016,Di-DeepLearningForTattoo-2017,Nicolas-WeightedAvgTattooIdentification-2022}. The main reasons that have slowed down the development of new tattoo retrieval methods are, on the one hand, the lack of large-scale tattoo databases, as shown in Tab.~\ref{tab:available_databases}. It should be noted that a few tattoo databases are publicly available and that most of them were designed for tattoo detection~\cite{Xu-NTUTattooV1-ICB-2016} and segmentation~\cite{Hrkac-Tattoo-DeMSI-2016,GonzalezSoler-SemiSyntheticTattoo-IWBF-2023} or provide a limited number of subjects~\cite{Nicolas-DeepGenericFeatures-CIARP-2019} or samples~\cite{Han-TattooImgSearch-PAMI-2019}. On the other hand, previous works often neglect challenges that cause performance degradation, \eg low-image quality, see Fig.~\ref{fig:tattoo_issues}. Consequently, a moderate identification rate (IR) below 65\% at the rank-1 is reported in the latest published tattoo identification pipelines.      

To overcome the aforementioned challenges, we introduce a large-scale semi-synthetic tattoo database based on which a novel tattoo retrieval approach is proposed that transforms tattooed human skin into the respective tattoo template to improve the final tattoo description. Thereby, the accuracy of tattoo retrieval can be significantly improved. The main contributions of this scientific work are:  

\begin{itemize}
    \item A large-scale semi-synthetic tattoo database that consists of 28,550 samples from 571 different tattoo templates or categories. To simulate a real-life scenarios, tattooed samples are generated and augmented in terms of scale, colour adjustments, distortions, and opacity. For the synthesis, the general requirements for generating synthetic biometric data are considered~\cite{Makrushin-GeneralRequiermentSyntehtic-2021}.   

    \item A tattoo retrieval solution that transforms tattooed human skin into the corresponding tattoo template. The reconstruction function optimised together with a well-known identity loss function alleviates the above issues encountered in real images and, thus, improves the identification performance of state-of-the-art algorithms.   

    \item A comprehensive evaluation of the proposed system on real images according to the metrics defined in the international standard ISO/IEC 19795-1~\cite{ISO-IEC-19795-1-Framework-210216} for biometric testing and reporting. Experimental results conducted on WebTattoo~\cite{Han-TattooImgSearch-PAMI-2019} and BIVTatt~\cite{Nicolas-DeepGenericFeatures-CIARP-2019} show a convincing performance improvement of the proposed system over the baselines.  
\end{itemize}

The remainder of this paper is organised as follows: a brief review of existing tattoo retrieval systems is provided in Sect.~\ref{sec:related_work}. In Sect.~\ref{sec:proposed_approach}, the fundamentals of the proposed approach are presented, along with a description of the tattoo generation. The experimental setup is explained in Sect.~\ref{sec:exp_setup} and results as well as derived findings are discussed in Sect~\ref{sec:results}. Finally, conclusions and future work directions are presented in Sect.~\ref{sec:conclusions}.

\section{Related Work}
\label{sec:related_work}

\begin{figure*}[!t]
    \centering
        \includegraphics[width=0.8\linewidth]{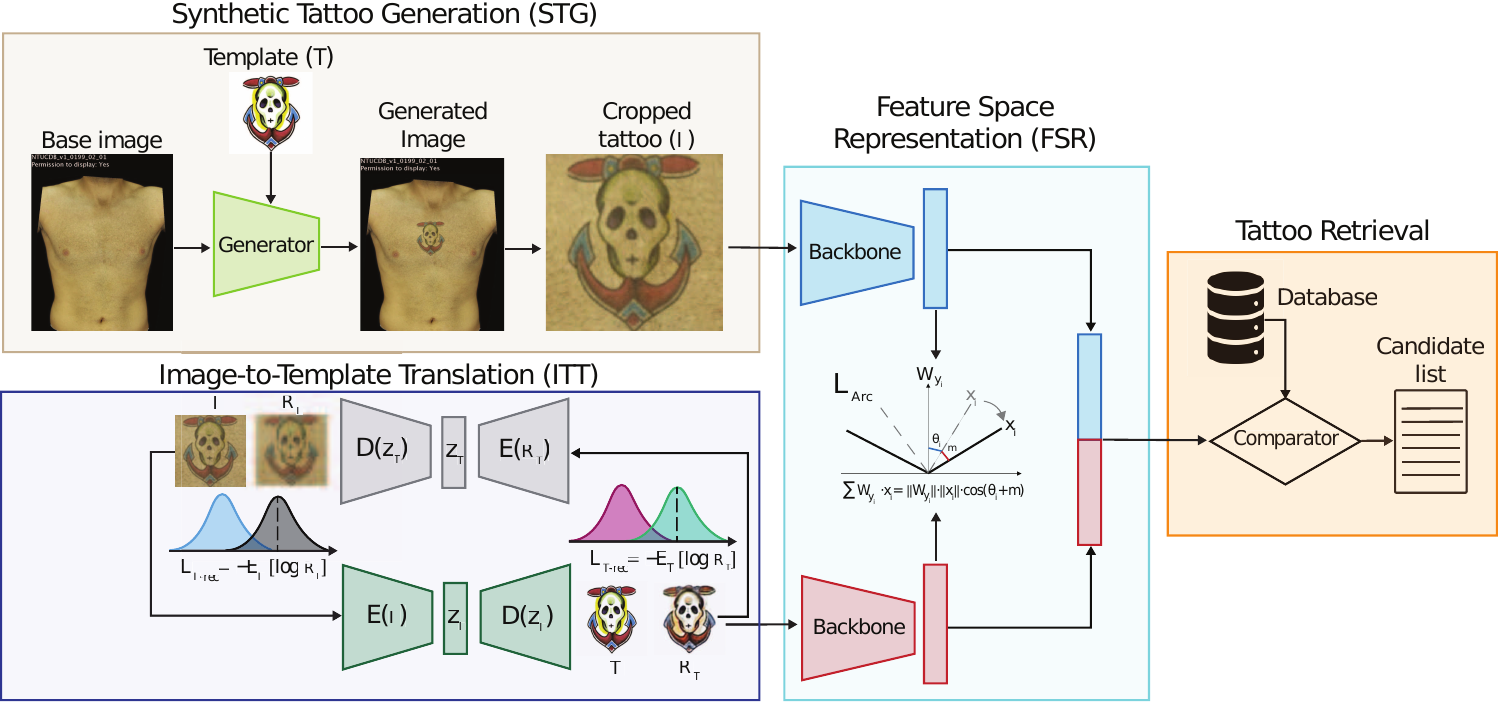}
    \caption{Conceptual overview of the proposed system: the template transformed from the input image helps to mitigate challenges related to the capturing process. The final feature embedding representing the salient properties of the input tattoo is the concatenation of the two computed feature embeddings and can be used to retrieve similar tattoo samples in a database.}
    \label{fig:overview}
\end{figure*}

Among the various soft biometric traits, tattoos have received considerable attention by forensic investigators in recent years, due to their prevalence among the criminal sector of the population and their prominence in visual attention. For more than 5,000 years, humans have marked their bodies with tattoos to express personal beliefs or to associate themselves with a group. The oldest evidence of tattooing was found on the body of Ötzi, the Iceman. Ötzi, Europe's most famous mummy, was discovered by German hikers in the Alps in 1991. The next oldest evidence of tattooing comes from mummies believed to have died between 3351 and 3017 BC in Ancient Egypt\footnote{\url{https://www.nationalgeographic.com/history/article/tattoos-mummies-ancient-cultures-symbols-meaning}}. In forensics, tattoos have also proved useful in helping to identify victims of terrorist attacks such as 9/11 and natural disasters such as the 2004 Indian Ocean tsunami~\cite{Jain-TattooID-PCM-2007}.

To assist in the identification of subjects, early tattoo retrieval systems relied on keyword or metadata matching. For this purpose, law enforcement agencies usually followed the ANSI/NIST-ITL 1-2000 standard and assigned a single keyword to each tattoo image in the database~\cite{McCabe-NISTTattoo-2000}. Since these solutions had obvious drawbacks, for example, $i)$ a limited vocabulary to describe numerous tattoo patterns, $ii)$ several keywords could be used to adequately describe a tattoo, and $iii)$ inconsistency in the labelling of tattoos, the next set of retrieval approaches focused on the use of handcrafted descriptors to represent the colour, texture, shape, and appearance of tattoos~\cite{Li-SIFTImageRetrieval-2009,Lee-ImageRetrievalForensics-2011,Han-TattooIdentSketch-ICB-2013}. A comprehensive review of these methods up to 2019 can be found in~\cite{Han-TattooImgSearch-PAMI-2019}.    

In this work, we restrict to briefly summarising deep learning-based tattoo retrieval solutions due to their reported success and outstanding performance in various pattern recognition and computer vision tasks~\cite{Joshi-SyntheticHumanAnalysis-PAMI-2024}. In 2019, Han \etal~\cite{Han-TattooImgSearch-PAMI-2019} introduced a system that was able to learn tattoo detection and compact tattoo representation jointly in a single DNN by multitask learning. Following this idea, Zhang \etal~\cite{Zhang-TaskIntegratedNetworks-PAMI-2020} proposed a DNN-based framework for joint tattoo detection and re-identification of individuals. Their reported IRs improved in the WebTattoo database~\cite{Han-TattooImgSearch-PAMI-2019} in case the detection module was activated. Nicol{\'a}s-D{\'\i}az \etal~\cite{Nicolas-DeepGenericFeatures-CIARP-2019} evaluated publicly available DNN models, pre-trained with large generic image databases, for tattoo identification and showed that these DNNs can achieve high IRs without even fine-tuning them for the target task of tattoo retrieval. The same authors also made the BIVTatt database publicly available in~\cite{Nicolas-DeepGenericFeatures-CIARP-2019}. In addition, Nicol{\'a}s-D{\'\i}az \etal~\cite{Nicolas-WeightedAvgTattooIdentification-2022} presented an attention pooling mechanism to adapt intermediate convolutional layers of pre-trained DNNs for the tattoo identification task.

The aforementioned tattoo retrieval systems are based on deep learning techniques, which require a considerable amount of training data to achieve good performance. In Tab.~\ref{tab:available_databases}, we summarise the main features of the databases used for such purposes. Note that, on the one hand, that Tatt-C and PinTatt are not available to the research community. On the other hand, the existing databases only consist of a few tattoo categories or samples which do not cover most real-life scenarios. This could make algorithms trained on those databases prone to overfitting. 

Image synthesis has recently emerged as a reliable solution to both privacy concerns and the lack of training data~\cite{Joshi-SyntheticHumanAnalysis-PAMI-2024}. To our knowledge, few studies have addressed the (semi-)synthetic generation of tattoo images for retrieval purposes. Existing methods to create face images with tattoos~\cite{Ibsen-FaceBeneathTheInk-AppliedSciences-2022}, are mainly designed for the visualisation of tattoos in controlled skin images~\cite{Calmon-AugmentedTattoo-2015}, or tattoo segmentation~\cite{GonzalezSoler-SemiSyntheticTattoo-IWBF-2023}. 

\section{Tattoo Template Reconstruction Network}
\label{sec:proposed_approach}

To mitigate the aforementioned problems, we introduce our Tattoo Template Reconstruction Network (TattTRN), which learns to map the input tattoo sample to its respective tattoo template. This tattoo template mapping is then used to enhance the distinctive attributes of the final feature embedding. Fig.~\ref{fig:overview} shows the conceptual overview of the proposed TattTRN. In our work, we hypothesise that a clean tattoo template may contain meaningful information to assist both an supervised learning approach and forensic investigators in making decisions. In the case of low-quality images, such as those shown in Fig.~\ref{fig:tattoo_issues}, where the tattoo is almost imperceptible to the human eye, the proposed concept could reconstruct the corresponding template and, thus, help forensic investigators. 
In addition to the above benefits, the transformation of an input tattoo image into a clean template would also provide subject identity suppression, as the reconstructed template is not expected not contain traces of skin colour or other sensible information.  

The following subsections describe the different components of our proposed TattTRN (Sect.~\ref{sec:approach}), as well as the definition and synthesis of the tattoo database (Sect.~\ref{sec:data_generation}) involved in training the approach.

\subsection{Main Components}
\label{sec:approach}

The proposed approach consists of three main components (see Fig.~\ref{fig:overview}): synthetic tattoo generation (STG), image-to-template translation (ITT) and feature space representation (FSR), which can be used for tattoo retrieval. STG allows the offline/online generation of tattooed samples, which can be directly represented as a feature embedding or translated into the predefined tattoo template. Whereas a direct representation of the sample (light blue trapezoid in FSR) aims at capturing intrinsic properties of the input image ($\mathcal{I}\in\mathbb{R}^{n\times n}$), the ITT module focuses on the construction of the respective clean tattoo template ($\mathcal{R}_{\mathcal{T}}$), which is, in turn, represented as a feature embedding (light red trapezoid in FSR). Both feature embeddings contain prominent characteristics of the input image and complement each other to enrich the final representation of the tattoos. For tattoo retrieval (a use case represented by the yellow box on the right in Fig.~\ref{fig:overview}), we use the concatenation of both embeddings to form a feature vector of size $2 \cdot K$. To translate the input image into a clean tattoo template in the ITT component, we use, as a proof-of-concept, the Unet~\cite{Ronneberger-UNet-MICCAI-2015} network based on the ResNet34 encoder~\cite{He-ResNet-CVPR-2016} and built a cyclic translation, as done in CycleGAN~\cite{Zhu-CycleGAN-CVPR-2017}. Thus, the quality of the template reconstruction is improved. Note that other encoder-decoder architectures proposed for image-to-image translation, such as Diffusion Models~\cite{Li-BBDMImage2Image-CVPR-2023}, could be employed, and thus improve the reconstruction results yielded by Unet. To optimise the cyclic Unet, two binary cross-entropy (BCE) loss functions are computed and combined as follows:
\begin{align}
     \mathcal{L}_{\mathcal{T}-rec} = & -\mathbf{E}_{\mathcal{T}}[\log~\mathcal{R}_{\mathcal{T}}] \nonumber\\
		                                    =  & -[\mathcal{T}\cdot \log~\mathcal{R}_{\mathcal{T}} + (1 - \mathcal{T})\cdot \log (1 - \mathcal{R}_{\mathcal{T}})],
    \label{eq:bce_loss_to_template}
\end{align}
\begin{align}
     \mathcal{L}_{\mathcal{I}-rec} = & -\mathbf{E}_{\mathcal{I}}[\log~\mathcal{R}_{\mathcal{I}}] \nonumber\\
		                                    =  & -[\mathcal{T}\cdot \log~\mathcal{R}_{\mathcal{I}} + (1 - \mathcal{T})\cdot \log (1 - \mathcal{R}_{\mathcal{I}})],
    \label{eq:bce_loss_to_img}
\end{align}
\begin{equation}
     \mathcal{L}_{rec} = \mathcal{L}_{\mathcal{T}-rec} + \mathcal{L}_{\mathcal{I}-rec} 
    \label{eq:bce_loss}
\end{equation}
    
\noindent where $\mathcal{I}$ is the semi-synthetic generated image using the tattoo template $\mathcal{T}$ in STG and $\mathcal{R}_{\mathcal{I}}$ is the reconstructed image from the translated tattoo template $\mathcal{R}_{\mathcal{T}}$ in ITT. BCE measures the difference between two probability distributions, \ie, ($\mathcal{T}$, $\mathcal{R}_{\mathcal{T}}$) and ($\mathcal{I}$, $\mathcal{R}_{\mathcal{I}}$) in this case and shows advantages in terms of convergence for image reconstruction over other loss functions such as the Minimum Squared Error (MSE). 

On top of the encoder-decoder network, a backbone network computes the embedding representation (light red trapezoid in FSR, Fig.~\ref{fig:overview}) of the reconstructed template $\mathcal{R}_{\mathcal{T}}$. In our work, both backbones (light blue and red trapezoids of FSR) are based on the same architecture and compute a feature embedding of size $K$ each. However, they do not share weights, as they process different representations of the input tattoo. Whereas the upper backbone (light blue trapezoid) of FSR is fed by the semi-synthetic raw tattoo image generated by the STG component, the lower backbone (light red trapezoid) computes the feature embedding of the clean tattoo template. 

Together with $\mathcal{L}_{rec}$ we select the ArcFace loss~\cite{Deng-ArcFace-CVPR-2019} to optimise the backbones separately. Angular margin penalty-based softmax loss was initially proposed to optimise face recognition systems~\cite{Deng-ArcFace-CVPR-2019,Boutros-ElasticFace-CVPR-2022}. It aims to extend the softmax decision boundary to improve intra- and inter-class variation by applying an angular penalty margin on the angle between feature embeddings and their respective weights. The ArcFace loss has reported high performance in the main benchmarks and is defined as follows~\cite{Deng-ArcFace-CVPR-2019}:
\begin{equation}
    \mathcal{L}_{Arc} = - \frac{1}{N}\sum_{i\in N} \log \frac{e^{s \cdot \cos(\theta_{y_i} + m)}}{e^{s \cdot \cos(\theta_{y_i} + m)} + \displaystyle\sum_{j=1, j \neq y_i}^{C} e^{s \cdot \cos \theta_j}},
    \label{eq:arc_loss}
\end{equation}

\noindent where $N$ and $C$ represent the training batch size and the number of classes (\ie, tattoo categories in our article), respectively, $y_i \in \{1, C\} \in \mathbb{N}$ is the tattoo category of the $i^{th}$ sample and $\theta_{y_i}$ is the angle between the tattoo feature embedding $x_i$ and the respective weight $w_{y_i}$. Note that linear function $x_i \cdot w_{y_i}$\footnote{The bias parameter of the fully connected layer representing the feature embedding is set to 0} can be represented as $||x_i||\cdot ||w_{y_i}|| \cdot \cos~\theta_{y_i}$. In our work, both $x_i$ and $w_{y_i}$ are normalised and hence  $||x_i|| = 1$ and $||w_{y_i}|| = 1$, resulting only in the softmax optimisation of $\cos~\theta_{y_i}$ along with the angular margin $m$ \ie, $\cos(\theta_{y_i} + m)$. $s$ is the scale factor according to~\cite{Wang-CosFace-CVPR-2018}.              

Finally, the loss combining both objective functions (\ie, Eq.~\ref{eq:bce_loss} and Eq.~\ref{eq:arc_loss}) to optimise our proposed TattTRN approach is defined as:
\begin{equation}
    \mathcal{L} = \frac{\mathcal{L}_{\mathcal{I}-Arc} + \mathcal{L}_{\mathcal{T}-Arc} + \lambda \cdot \mathcal{L}_{rec}}{3},
\end{equation}
\noindent where the hyperparameter $\lambda$ was empirically set to 4 and aims to keep the independently produced losses from both objective functions in equilibrium. $\mathcal{L}_{\mathcal{I}-Arc}$ and $\mathcal{L}_{\mathcal{T}-Arc}$ represent the ArcFace losses for the backbones optimised over $\mathcal{I}$ and $\mathcal{R}_{\mathcal{T}}$, respectively.  
    
\subsection{Tattoo Generation}
\label{sec:data_generation}

\begin{figure}[!t]
    \centering
    \begin{subfigure}{0.38\linewidth}
        \includegraphics[width=\textwidth]{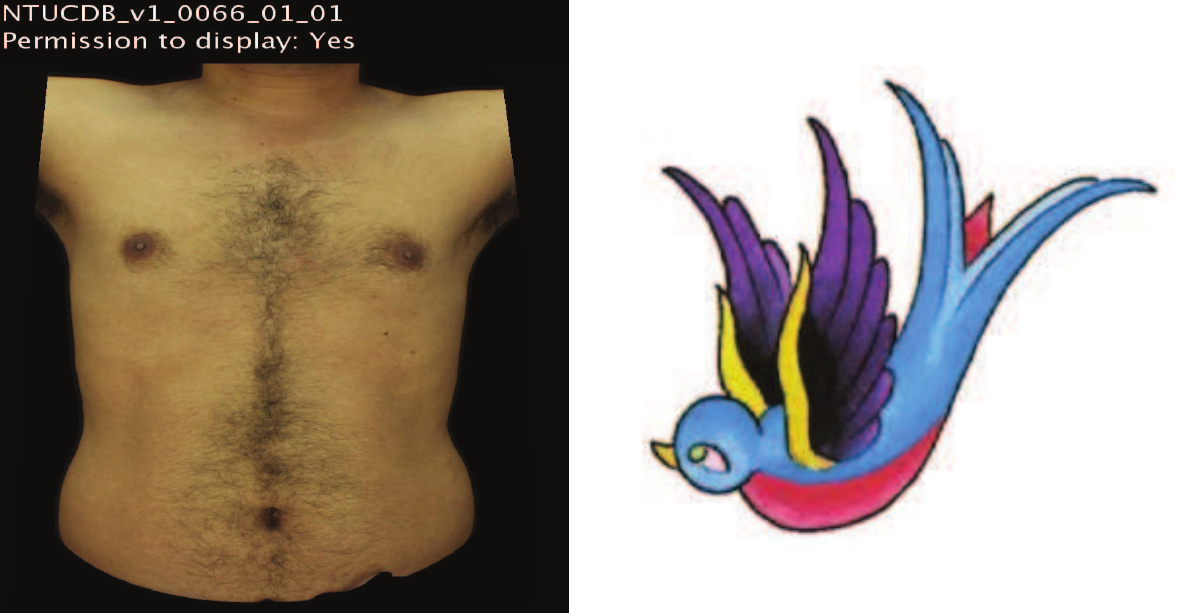}
    \end{subfigure}
    \begin{subfigure}{0.38\linewidth}
        \includegraphics[width=\textwidth]{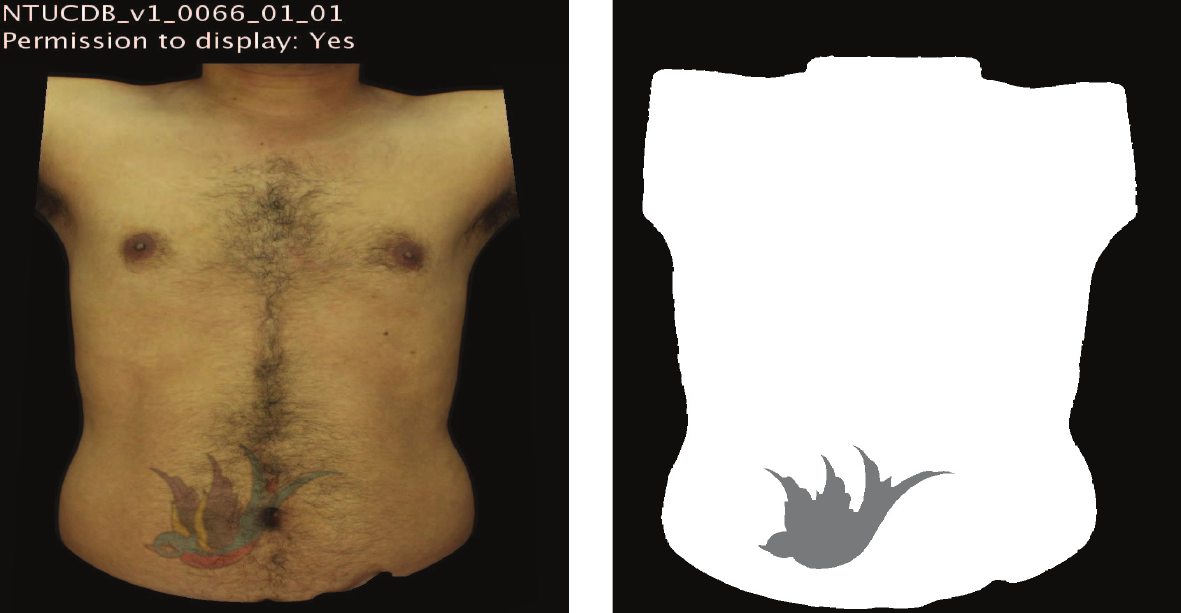}
    \end{subfigure}
    \begin{subfigure}{0.183\linewidth}
        \includegraphics[width=\textwidth]{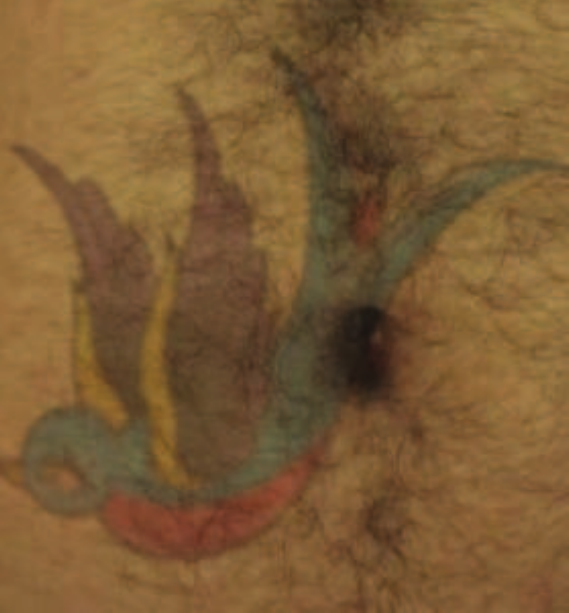}
    \end{subfigure}\\
    \begin{subfigure}{0.38\linewidth}
        \includegraphics[width=\textwidth]{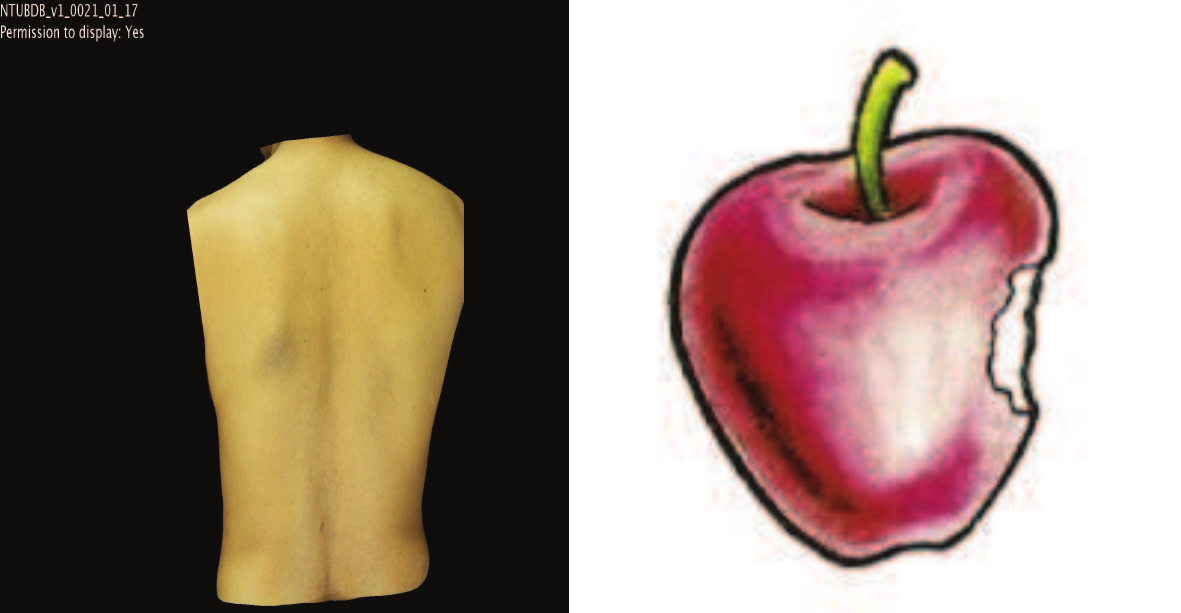}
        \caption{}
        \label{fig:back_orig}
    \end{subfigure}
    \begin{subfigure}{0.38\linewidth}
        \includegraphics[width=\textwidth]{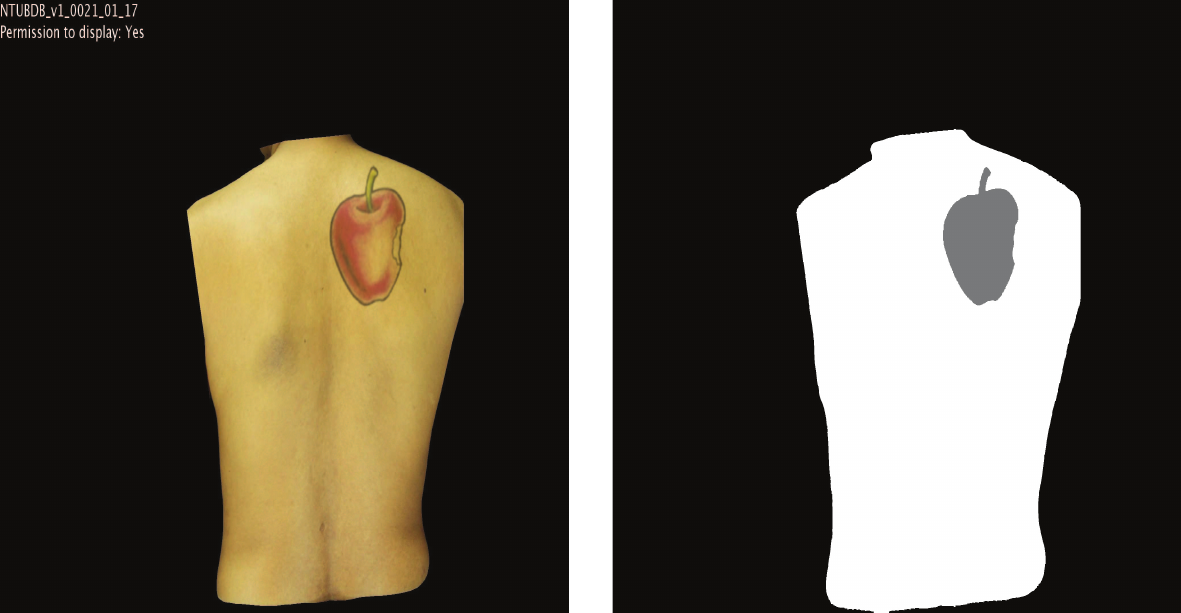}
        \caption{}
        \label{fig:back_generated}
    \end{subfigure}
    \begin{subfigure}{0.183\linewidth}
        \includegraphics[width=\textwidth]{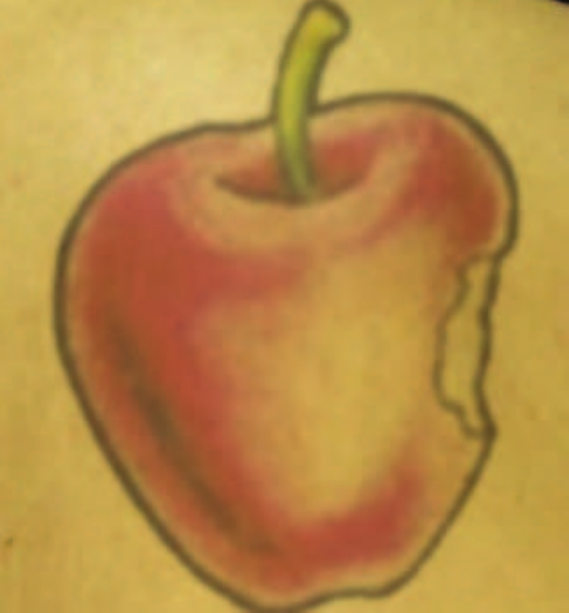}
        \caption{}
        \label{fig:cropped_back}
    \end{subfigure}
    \caption{Examples of tattoos generated on chest and back images with their respective segmentation maps (\ref{fig:back_generated}) using base images and random tattoo templates (\ref{fig:back_orig}). Cropped tattoo images used in the network training (\ref{fig:cropped_back}).}
    \label{fig:tattoo_generation}
\end{figure}

Synthetic image generation has mainly focused on facial characteristics. In terms of market, face recognition has maintained a stable growth and its use has spread to many application contexts, such as financial transactions, border control and video surveillance. To synthesise high-quality facial images, several approaches based on GAN or diffusion models, such as the StyleGAN family~\cite{Karras-StyleGAN2-2020,Karras-StyleGAN3-2021} and latent diffusion models~\cite{Rombach-LatentDiffModels-CVPR-2022}, have been proposed in recent years. In comparison to the face, tattoo synthesis has received little attention. Recently, Andrej Karpathy demonstrated the robustness of diffusion models for generating realistic tattooed human images\footnote{\url{https://www.youtube.com/watch?v=sM9bozW295Q}}. Ibsen \etal~\cite{Ibsen-FaceBeneathInk-2022} proposed a tattoo generator to blend predefined tattoo templates with real faces and evaluated the performance impact of facial recognition systems on tattooed faces. Following this idea, Gonzalez-Soler \etal~\cite{GonzalezSoler-SemiSyntheticTattoo-IWBF-2023,GonzalezSoler-ImpactTattooedHands-BIOSIG-2023} extended the previous pipeline to mix the predefined tattoo templates with any area of human skin and demonstrated its utility for the segmentation of real tattoos. In this article, we reuse the ideas shown in~\cite{GonzalezSoler-SemiSyntheticTattoo-IWBF-2023,GonzalezSoler-ImpactTattooedHands-BIOSIG-2023} to generate 28,550 images of 571 different tattoo templates, \ie, 50 images per template. The NTU-Back~\cite{Nurhudatiana-AutIdentPigmented-2013automated,Nurhudatiana-IndividualitySkin-TIFS-2013} and NTU-Chest~\cite{Nurhudatiana-OnCriminalIdSkin-TIFS-2015,Nurhudatiana-IndividualitySkin-TIFS-2013} databases, consisting of 647 and 434 images respectively, are used as base images of human skin to blend with the tattoos. Randomly 25:25 images from both databases are used as base images to synthesise the 50 images per tattoo template. Fig.~\ref{fig:tattoo_generation} shows examples of tattooed human skins. To simulate real images, the tattoo is made more realistic by adjusting the colour, and Gaussian blur, and reducing the opacity. Since the segmentation map is available, the final tattooed samples are then cropped to train the TattTRN system.             

\section{Experimental Setup}
\label{sec:exp_setup}

The experimental evaluation goals are manifold: $i)$ study the utility of the proposed semi-synthetic database for training a tattoo retrieval system capable of retrieving real tattoo samples, $ii)$ establish a benchmark of the proposed TattTRN systems with baseline approaches, and $iii)$ evaluate the capability of the TTE subsystem to enhance low-quality images. In all experiments, we follow a cross-database protocol where the proposed semi-synthetic database is used to train the systems and the real ones for evaluating performance. The test set is randomly divided into five subsets of biometric enrolment and identification transactions following a closed-set scenario, \ie, searched identities are always in the enrolment database. Therefore, the mean identification rates alongside the standard deviation are presented as Cumulative Matching Characteristic (CMC) curves. Formally, CMC is a graphical presentation of the results of mated searches in a closed-set identification test, which plots the true positive identification rate (IR) as a function of a rank value~\cite{ISO-IEC-19795-1-Framework-210216}. The identification performance of the systems is also evaluated for an open set scenario in terms of False Negative Identification Rates (FNIR) and False Positive Identification Rates (FPIR), and their values are presented as DET curves\cite{ISO-IEC-19795-1-Framework-210216}.          

\begin{figure}[!t]
    \centering
    \begin{subfigure}{\linewidth}
        \includegraphics[width=\textwidth]{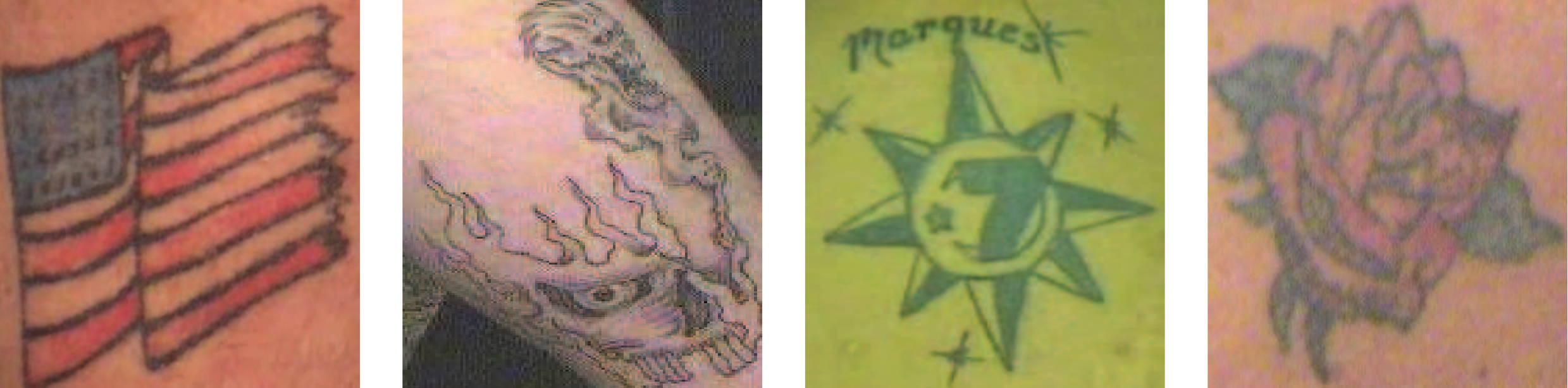}
        \caption{WebTattoo}
        \label{fig:webtattoo}
    \end{subfigure}
    \begin{subfigure}{\linewidth}
        \includegraphics[width=\textwidth]{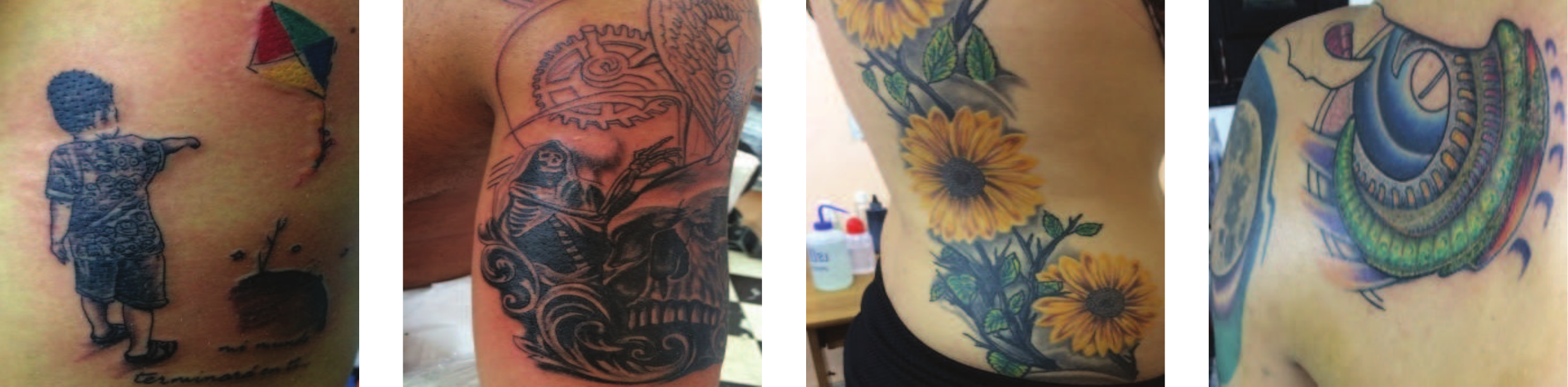}
        \caption{BIVTatt}
        \label{fig:bivtatt}
    \end{subfigure}
    \caption{Examples of the database used to evaluate the proposed TattTRN system.}
    \label{fig:database_examples}
\end{figure}

\begin{table*}[!t]
\centering
\caption{Closed-set identification performance of the proposed TattTRN approach for different backbones in terms of Rank-1 (\%) on real images in WebTattoo~\cite{Han-TattooImgSearch-PAMI-2019} and BiVTatt~\cite{Nicolas-DeepGenericFeatures-CIARP-2019}. The best result per backbone and database is highlighted in bold.}
\label{tab:parameter_op}
\begin{adjustbox}{width=\linewidth,center}
    \begin{tabular}{r| c c c| c c c| c c c| c c c| c c c| c c c} \toprule
\textbf{Database} &    \multicolumn{9}{c|}{\textbf{WebTattoo}}    &  \multicolumn{9}{c}{\textbf{BIVTatt}}   \\
                          $m$ & \multicolumn{3}{c|}{\textbf{0.1}} & \multicolumn{3}{c|}{\textbf{0.5}} & \multicolumn{3}{c|}{\textbf{0.9}} & \multicolumn{3}{c|}{\textbf{0.1}} & \multicolumn{3}{c|}{\textbf{0.5}} & \multicolumn{3}{c}{\textbf{0.9}} \\
     \textbf{Backbone}/$K$ &   \textbf{128}    &   \textbf{256}  &  \textbf{512}  &   \textbf{128}   &  \textbf{256}  &  \textbf{512}    &   \textbf{128}   &  \textbf{256}  &  \textbf{512}  &   \textbf{128}    &   \textbf{256}  &  \textbf{512}  &   \textbf{128}   &  \textbf{256}  &  \textbf{512}    &   \textbf{128}   &  \textbf{256}  &  \textbf{512} \\ 
    \midrule
     MobileNetv3      & 70.98 & 73.27 & 76.17 & 73.27 & 75.58 & \textbf{76.51} & 66.39 & 71.89 & 63.73 & 94.40 & 93.58 & 94.40 & 94.21 & 94.40 & \textbf{94.65} & 91.95 & 92.33 & 89.25    \\
     ResNet101        & 70.47 & 73.84 & 78.13 & 74.25 & 76.63 & \textbf{78.45} & 60.52 & 63.17 & 63.46 & 93.46 & 94.09 & 94.59 & 93.65 & \textbf{94.78} & 94.47 & 88.43 & 88.62 & 87.92       \\
     DenseNet121      & 73.59 & \textbf{78.67} & 76.78 & 74.64 & 77.27 & 78.13 & 75.01 & 77.03 & 76.03 & 93.14 & 95.16 & 93.96 & 93.33 & 94.28  & 94.72 & 94.72 & 94.53 & \textbf{95.22}  \\
     EfficientNetv2   & 79.43 & 78.92 & \textbf{80.44} & 79.19 & 79.61 & 79.63 & 71.99 & 59.12 & 62.04 & 94.47 & 95.41 & 96.29 & 95.41 & \textbf{96.54} & 96.16 & 93.08 & 82.58 & 86.73     \\
     Swin             & 77.76 & 78.80 & \textbf{81.60} & 77.84 & 78.82 & 80.57  & 69.26 & 66.51 & 69.88 & 95.35 & 94.47 & \textbf{96.54} & 93.96 & 94.47 & 95.91 & 89.37 & 85.91 & 85.41 \\
     \midrule
     Avg.             & 74.45 & 76.70 & 78.62 & 75.84 & 77.58 & 78.66 & 68.63 & 67.54 & 67.03 & 94.16 & 94.54 & 95.16 & 94.11 & 94.89 & 95.18 & 91.51 & 88.79 & 88.91\\
     
\bottomrule

    \end{tabular}
\end{adjustbox}
\end{table*} 

\subsection{Databases}

To evaluate the utility of both the semi-synthetic database and the proposed system, we selected the two publicly available databases in Tab.~\ref{tab:available_databases} that provide tattoo category labels \ie, WebTattoo~\cite{Han-TattooImgSearch-PAMI-2019} and BIVTatt~\cite{Nicolas-DeepGenericFeatures-CIARP-2019}.

According to~\cite{Han-TattooImgSearch-PAMI-2019}, WebTattoo comprises around 300K samples of 600 different tattoo categories which were extracted from the internet. For training, the authors randomly selected about 1,400 tattoo images from 400 tattoo classes and the remaining tattoo images from 200 tattoo classes were used for testing. In our experiments, we only used the training set that was made available to the research community.

BIVTatt~\cite{Nicolas-DeepGenericFeatures-CIARP-2019} contains 210 original tattoo images and 4,200 images generated after applying 20 different types of transformations to the original images. Along with the images, the authors provided the bounding box information for each tattoo. Given the coordinate problems associated with the bounding boxes, only correctly cropped samples were selected, resulting in 3,103 samples from 159 tattoo categories. Fig.~\ref{fig:database_examples} shows examples of tattoo images in the WebTattoo~\cite{Han-TattooImgSearch-PAMI-2019} and BIVTatt~\cite{Nicolas-DeepGenericFeatures-CIARP-2019} databases.  

\subsection{Implementation Details}

Both the proposed TattTRN systems and baselines were implemented in PyTorch~\cite{Paszke-PyTorchAnImperative-2019} and trained utilising a Nvidia A100 Tensor Core GPU with 40 GB of GPU Memory over the generated 28,550 semi-synthetic images. The image size was set to 224~$\times$~224. To further cover the feature space of real tattoo images, the brightness, contrast, saturation, and hue of the training images are randomly adjusted. In addition, the networks were initialised with their pre-trained weights on ImageNet~\cite{Deng-ImageNet-CVPR0-2009} and trained for 100 epochs using the Adam optimiser with a learning rate of $1\mathrm{e}{-5}$ and weight decay of 0.95. A batch size of 64 images is also set for training. As backbones, the large version of MobileNetv3~\cite{Howard-MobileNetv3-CVPR-2019}, 101-layer ResNet~\cite{He-ResNet-CVPR-2016}, 121-layer DenseNet~\cite{Huang-DenseNet-CVPR-2017}, and the small version of EfficientNetv2~\cite{Tan-EfficientNetv2-ICML-2021} and SwinTransformer (Swin)~\cite{Liu-SwinTransformer-CVPR-2021} are used. These networks offer coverage of the main approaches proposed in the last decade \eg, attention mechanisms~\cite{Hu-SqueezeAndExcitation-CVPR-2018} and vision transformers~\cite{Vaswani-AttentionVisualTransformers-2017attention}. Note that other architectures developed specifically for tattoo retrieval \eg, the joint detection and compact representation scheme~\cite{Han-TattooImgSearch-PAMI-2019} or the weighted average pooling-based approach~\cite{Nicolas-WeightedAvgTattooIdentification-2022}\footnote{The implementation of both approaches has not been made publicly available.}, can also be applied.      

\section{Results and Discussion}
\label{sec:results}

The result's discussion relies on the above three goals: the closed-set identification performance of the proposed TattTRN approach (Sect.~\ref{sec:closed_set}), as well as the respective open-set identification results~\ref{sec:open_set} are reported. Sect.~\ref{sec:benchmark} shows a comparison of TattTRN with the baselines. To evaluate the utility of TattTRN, the benchmark is established against the direct embedding representation of the semi-synthetic tattoo images generated by the STG component. Finally, some examples of translating real images into templates are presented in Sect.~\ref{sec:real_translation}.        

\subsection{Closed-set Evaluation}
\label{sec:closed_set}

\begin{figure}[!t]
    \centering
    \begin{subfigure}{0.8\linewidth}
        \includegraphics[width=\textwidth]{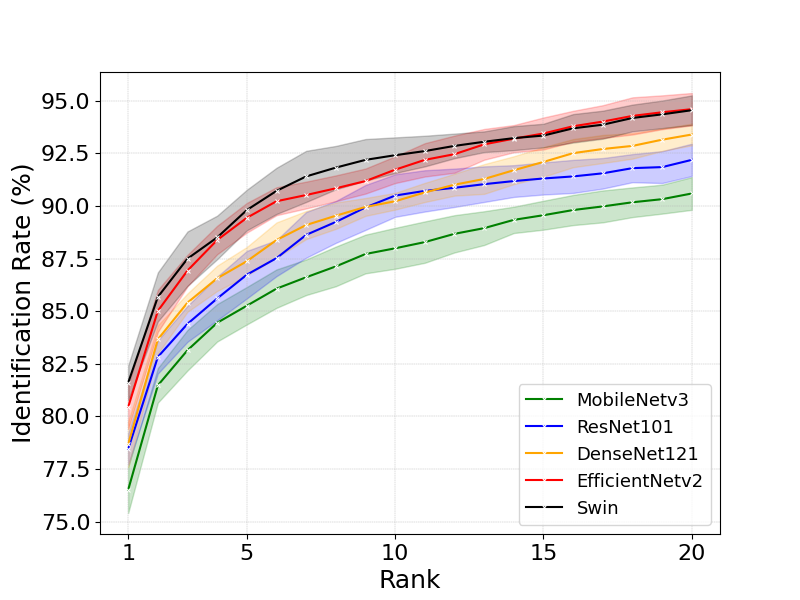}
        \caption{WebTattoo}
        \label{fig:webtattoo_cmc}
    \end{subfigure}
    \begin{subfigure}{0.8\linewidth}
        \includegraphics[width=\textwidth]{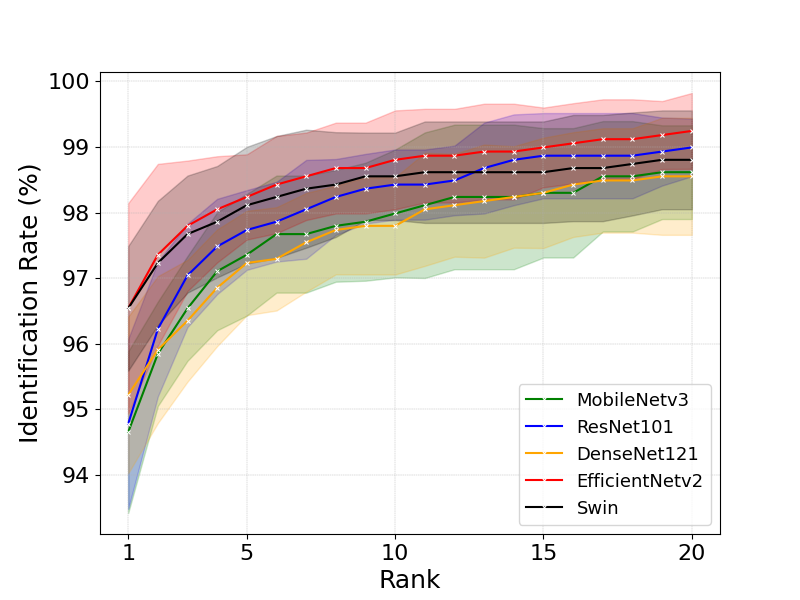}
        \caption{BIVTatt}
        \label{fig:bivtatt_cmc}
    \end{subfigure}
    \caption{CMC curves for different backbones combined with TattTRN.}
    \label{fig:cmc}
\end{figure}

\begin{figure}[!t]
    \centering
    \begin{subfigure}{0.49\linewidth}
        \includegraphics[width=\textwidth]{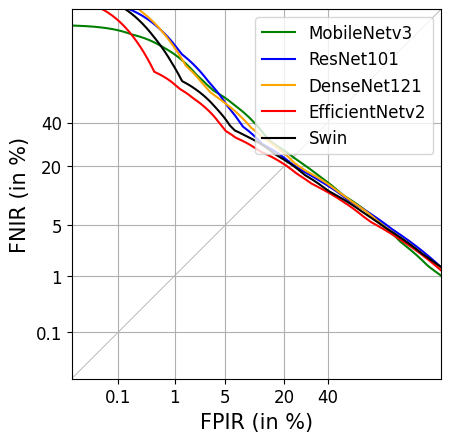}
        \caption{WebTattoo}
        \label{fig:webtattoo_det}
    \end{subfigure}
    \begin{subfigure}{0.49\linewidth}
        \includegraphics[width=\textwidth]{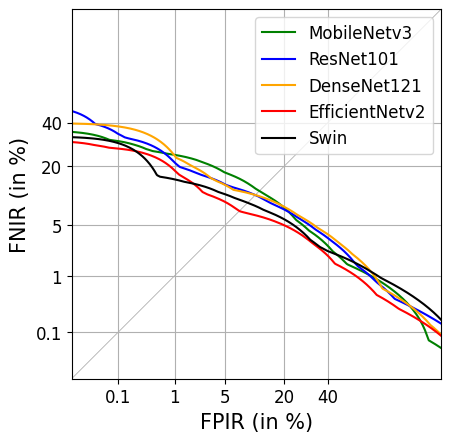}
        \caption{BIVTatt}
        \label{fig:bivtatt_det}
    \end{subfigure}
    \caption{DET curves for different backbones combined with TattTRN.}
    \label{fig:det}
\end{figure}

Since the use of ArcFace loss depends on the optimisation of the angular margin parameter (\ie, $m$), we report the mean closed-set identification performance of TattTRN for three values of $m = \{0.1, 0.5, 0.9\}$ and three embedding sizes $K = \{128, 256, 512\}$ in Tab.~\ref{tab:parameter_op}. $K$ values greater than 512 would result in a final feature vector of $2 \cdot K > 1024$, leading to an efficiency deterioration of the system.    

Note that the generated semi-synthetic tattoo images reflect the main properties of the real tattooed samples and their use allows our TattTRN approach to achieve high performance for the evaluated real databases. TattTRN is able to register an average IR of at most 81.60\% at rank-1 for the challenging WebTattoo dataset and 96.54\% at the same rank value for BIVTatt. Compared to the results reported in their respective articles, TattTRN reports a performance improvement of approximately 18\% for WebTattoo~(\ie, IR~$\approx$~64\%)~\cite{Han-TattooImgSearch-PAMI-2019} and 25\% for BIVTatt~(\ie, IR~=~70.91\%)~\cite{Nicolas-DeepGenericFeatures-CIARP-2019}. It is worth mentioning that the benchmarking systems referred to in the articles of both databases were fully trained on real tattoo samples following an internal database protocol, in contrast to TattTRN, representing the most salient properties of the semi-synthetic images in the training to identify real specimens. In terms of parameter optimisation, we observe that most backbones achieve on average the best performance for $K$ = 512 and $m$ = 0.5 or $m$ = 0.1, with the combination of TattTRN with Swin~\cite{Liu-SwinTransformer-CVPR-2021} being the best performing approach for WebTattoo (\ie, IR~=~81.60\%) and BIVTatt (\ie, IR~=~96.54\%) in rank-1.  

Since in forensic investigations not only the first positions in the candidate lists are important, we also show the CMC curves for the different backbones combined with the TattTRN approach proposed in Fig.~\ref{fig:cmc} using the best parameter configurations (\ie, bold values in Tab.~\ref{tab:parameter_op}). Note that TattTRN combined with Swin achieves only the best IR in rank-1 for both databases. For rank values above 15, we can see that the EfficientNetv2 backbone yields an average IR of about 95\% for WebTattoo and above 99\% for BIVTatt. This implies that forensic investigators can detect a perpetrator based on tattoos with an accuracy of up to 99\% by checking the first 20 entries of the candidate list.     

\subsection{Open-set Evaluation}
\label{sec:open_set}

We report the TattTRN performance in Fig.~\ref{fig:det} for the scenario where the tattoo template or category is probably not included in the enrolment set (\ie, open-set scenario). Note that TattTRN combined with different backbones yields similar equal error rates (EER) for WebTattoo. However, in line with the closed-set results shown in Sect.~\ref{sec:closed_set}, EfficientNetv2 achieves the best identification performance for high-security thresholds for both databases. In particular, this backbone achieves an FNIR of more than 70\% for WebTattoo and an FNIR of approximately 30\% for BIVTatt for an FPIR = 0.1\%. The latter result implies that, at most, 30 out of 100 mated transactions are not included in the candidate list if the system accepts only 1 out of 1000 non-mated transactions in the candidate list.

\subsection{Benchmark of TattTRN}
\label{sec:benchmark}

\begin{figure}
    \centering
    \includegraphics[width=0.8\linewidth]{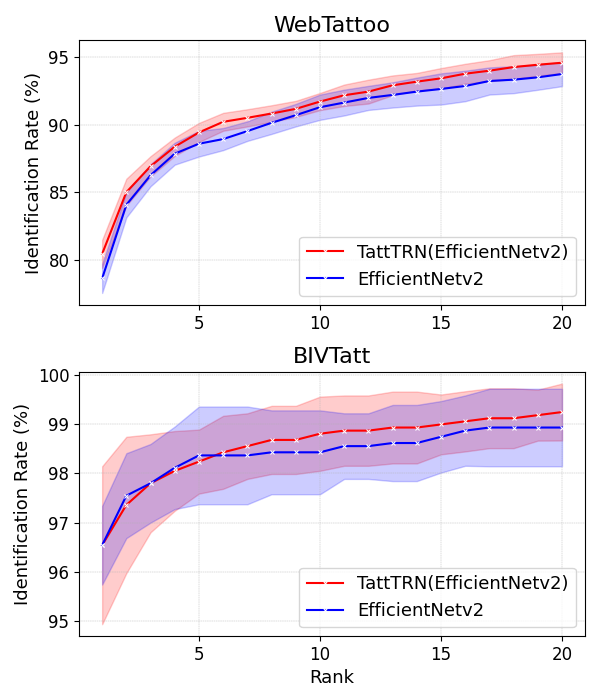}
    \caption{Benchmark of TattTRN against the respective best-performing backbone.}
    \label{fig:cmc_benchmark}
\end{figure}

\begin{figure*}[!t]
    \centering
    \begin{subfigure}{0.49\linewidth}
        \includegraphics[width=\textwidth]{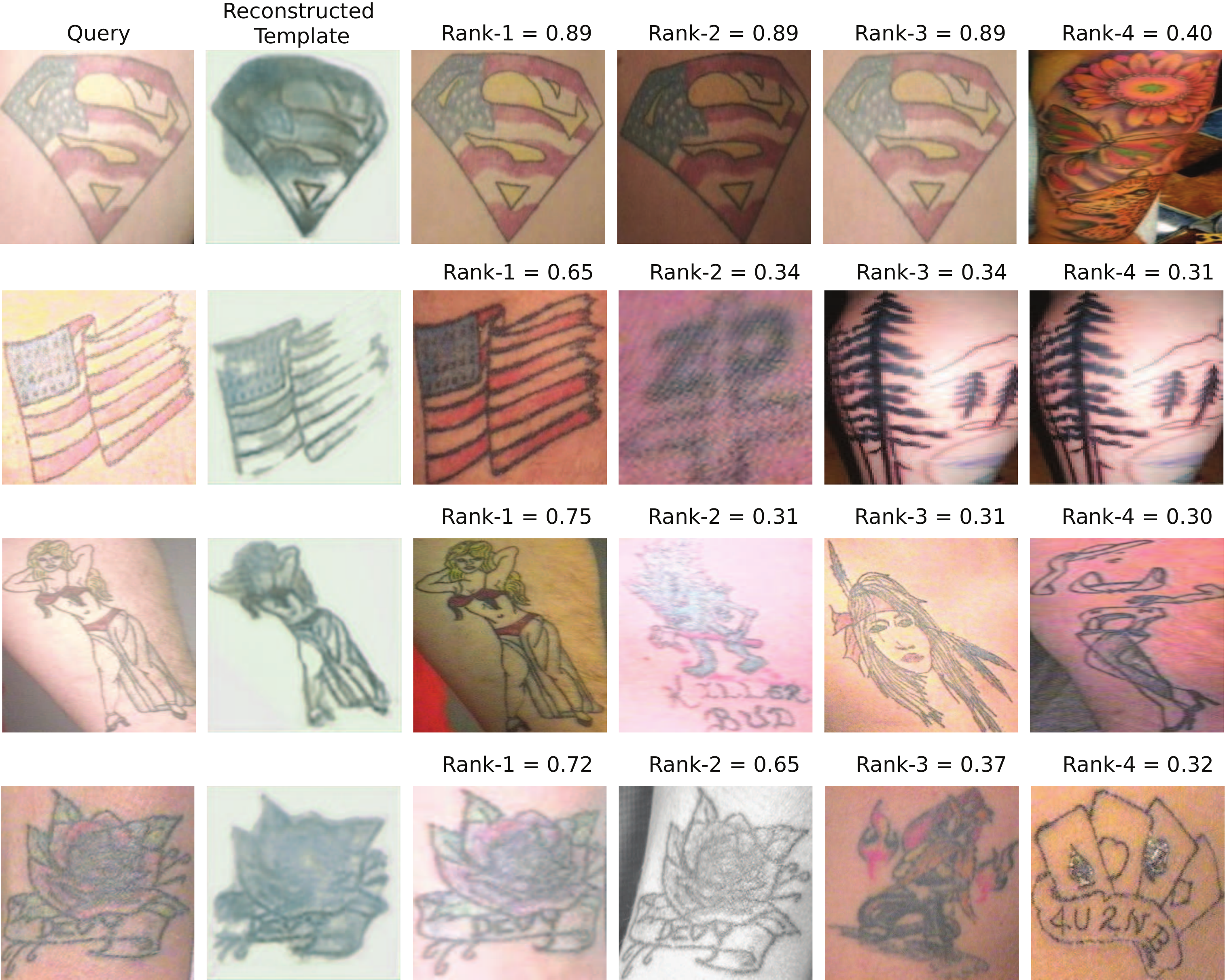}
        \caption{Good reconstructed templates.}
        \label{fig:good_samples}
    \end{subfigure}\hspace{0.2cm}
    \begin{subfigure}{0.49\linewidth}
        \includegraphics[width=\textwidth]{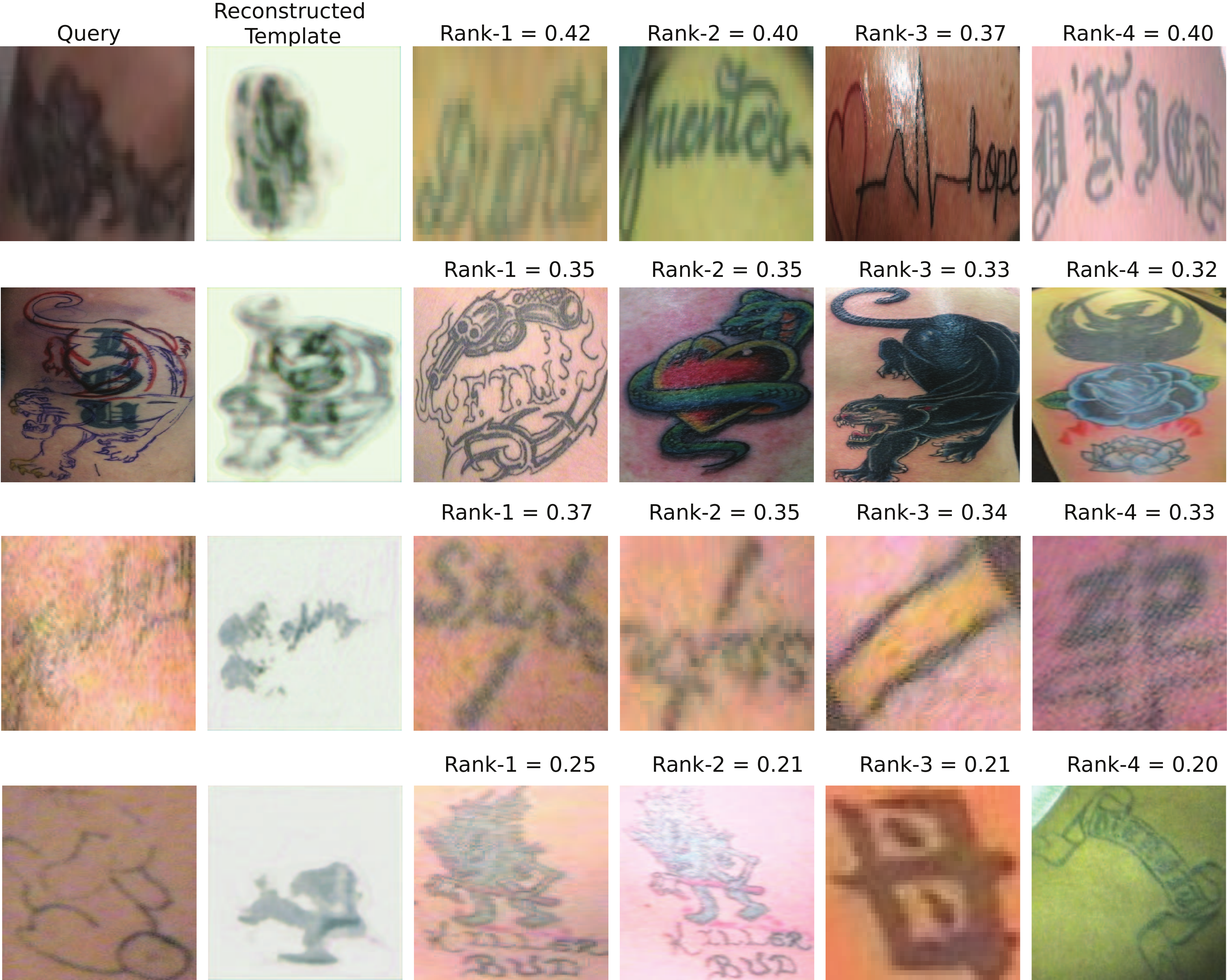}
        \caption{Bad reconstructed templates.}
        \label{fig:bad_samples}
    \end{subfigure}
    \caption{Example of correctly and wrongly retrieved tattooed samples from WebTattoo~\cite{Han-TattooImgSearch-PAMI-2019} for rank values between 1 and 4. The cosine similarity values are given for each case.}
    \label{fig:examples_reconstructed}
\end{figure*}

We evaluate the benefits of the translation module (\ie ITT) for TattTRN. For this purpose, the entire TattTRN architecture is compared with the submodule comprised of only the raw embedding representation of the generated semi-synthetic tattoo image (\ie light blue trapezoid in Fig.~\ref{fig:overview}). Fig.~\ref{fig:cmc_benchmark} shows the benchmark of the whole TattTRN against the raw embedding computed by EfficientNetv2~\cite{Tan-EfficientNetv2-ICML-2021}. Note that the TattTRN benefits from the ITT component, resulting in a performance improvement for EfficientNetv2 on both databases. Specifically, for the challenging WebTattoo dataset, TattTRN outperforms EfficientNetv2 for all rank values and achieves an IR approximate of 95\% in rank-20 \ie, the perpetrator can be identified with an accuracy of up to 95\% by checking at most the first 20 entries of the candidate list. Despite the advantages of the ITT component, there are still several images in WebTattoo whose quality should be further improved (see Sect.~\ref{sec:real_translation}). It should be mentioned that extreme sharpening or blurring transformations to decrease image quality, as well as approaches that cut out part of the image (\eg, CutMix~\cite{Yun-CutMix-CVPR-2019}), were not taken into account in the training. The use of these operations could also increase the reconstruction performance of IIT, and hence of TattTRN.                

\subsection{Tattoo Retrieval Examples}
\label{sec:real_translation}

Finally, in Fig.~\ref{fig:examples_reconstructed} we show some examples of candidate lists retrieved by TattTRN from searched images and the respective translated templates. Note that TattTRN is able to correctly build the template from those images that were successfully retrieved in the first positions of the candidate list (\ie, images in Fig.~\ref{fig:good_samples}). The ITT component is therefore capable of encoding challenging patterns, such as the woman and the rose in the last two rows of Fig.~\ref{fig:good_samples}. It should be noticed that the remaining retrieved images, ranked from 2 to 4 in rows 2 and 3 of Fig.~\ref{fig:good_samples}, do not belong to the consulted tattoo category: TattTRN reports a similarity below 0.35.    

Fig.~\ref{fig:bad_samples}, on the other hand, shows some images for which TattTRN could not correctly construct their clean tattoo template, resulting in lower similarity values for the first positions in the candidate list. Note that they either have extremely low image quality (\eg, images of rows 1 and 3), which might even be difficult for human analysis, or they contain overlapping tattoos (\eg, image of the second row). In the latter case, artistic characters are displayed in the background and an in-process panther in the foreground: TattTRN was able to partially construct its clean tattoo template. The overlapping of tattoos resulted in TattTRN regaining first place in the ranking for a tattoo containing Latin characters and third place for the corresponding correct tattoo category (black panther). Notice that all the tattoos retrieved in Fig.~\ref{fig:bad_samples} have a similarity of less than 0.42 with respect to the queried tattoo samples. This implies, in line with the results in Fig.~\ref{fig:webtattoo_det}, that the input image would be a false negative depending on the system threshold (typically 0.5) and therefore all items in the candidate list would be rejected by the system, even if there are items of the same category as the input image at the former positions in the candidate list (\eg, the black panther). A potential solution to improve the results of TattTRN on the above images would be based on the use of a prompt-guided approach that also encodes the semantic meaning of the tattoo and the overlap between tattoos, or the combination with a super-resolution method that further improves the image quality of the input tattoos.            

\section{Conclusions}
\label{sec:conclusions}

This work proposes a semi-synthetic tattoo database in conjunction with a tattoo retrieval approach called TattTRN, which exploits the transformation of the input image into a clean tattoo template to enrich the final feature embedding and thus improve the retrieval performance of difficult images in two freely available databases: WebTattoo and BIVTatt. The experimental evaluation of TattTRN showed that the proposed balanced database, consisting of 28,550 images from 571 tattoo categories, gathers the main properties of the real images, resulting in an IR of up to 99\% in the top 20 positions of the candidate list. Experimental results also reported high performance when TattTRN is combined with EfficientNetv2, yielding an IR of approximately 81\% for WebTattoo in rank-1. Compared to the results reported for this database in its corresponding article, TattTRN improves identification performance by 18 \% (\ie, 81.60\% vs. 64\%). Despite the results obtained by the TattTRN, it still fails to encode extremely low-quality images or images containing overlapping tattoos. In future work, we will combine TattTRN with a new prompt-based component that includes the semantic meaning of tattoos and evaluate the effect of different skin tones on tattoo segmentation and retrieval.         

\section{Acknowledgement}

This research work has been partially funded by the Hessian Ministry of the Interior and Sport in the course of the Bio4ensics project and the German Federal Ministry of Education and Research and the Hessian Ministry of Higher Education, Research, Science and the Arts within their joint support of the National Research Center for Applied Cybersecurity ATHENE.

{
    \small
    \bibliographystyle{ieeenat_fullname}
    \bibliography{arXiv}
}


\end{document}